\documentclass[10pt,letterpaper]{article}
\usepackage[top=0.85in,left=2.75in,footskip=0.75in]{geometry}

\usepackage{amsmath,amssymb}

\usepackage{changepage}

\usepackage[utf8x]{inputenc}

\usepackage{textcomp,marvosym}

\usepackage{cite}

\usepackage{nameref,hyperref}

\usepackage[right]{lineno}

\usepackage{microtype}
\DisableLigatures[f]{encoding = *, family = * }

\usepackage[table]{xcolor}

\usepackage{array}

\usepackage{diagbox}
\usepackage{makecell}
\newcolumntype{+}{!{\vrule width 2pt}}

\newlength\savedwidth

\raggedright
\usepackage[aboveskip=1pt,labelfont=bf,labelsep=period,justification=raggedright,singlelinecheck=off]{caption}

\makeatletter
\renewcommand{\@biblabel}[1]{\quad#1.}
\makeatother

\usepackage{lastpage,fancyhdr,graphicx}
\usepackage{epstopdf}
\fancyheadoffset[L]{2.25in}
\fancyfootoffset[L]{2.25in}
\begin{document}
\vspace*{0.2in}

\begin{flushleft}
{\Large
\textbf\newline{Learning brain MRI quality control: a multi-factorial generalization problem} 
}
\newline
\\
Ghiles Reguig\textsuperscript{1,2, *},
Marie Chupin\textsuperscript{1,3},
Hugo Dary\textsuperscript{1,3},
Eric Bardinet\textsuperscript{1},
Stéphane Lehéricy\textsuperscript{1,4},
Romain Valabregue\textsuperscript{1,3}
\\
\bigskip
\textbf{1} Paris Brain Institute – ICM, INSERM U 1127, CNRS UMR 7225, Sorbonne Université, Paris, France
\\
\textbf{2} Siemens Healthineers, Saint-Denis, France
\\
\textbf{3} CATI, US52-UAR2031, CEA, ICM, SU, CNRS, INSERM, APHP, Ile de France, France 

\textbf{4} Department of Neuroradiology, Pitié-Salpêtrière Hospital, Public Assistance - Paris Hospitals (AP-HP), Paris, France
\\
\bigskip

%
%





* contact: ghiles.reguig@gmail.com

\end{flushleft}
\section*{Abstract}
Due to the growing number of MRI data, automated quality control (QC) has become essential, especially for larger scale analysis. Several attempts have been made in order to develop reliable and scalable QC pipelines. However, the generalization of these methods on new data independent of those used for learning is a difficult problem because of the biases inherent in MRI data. This work aimed at evaluating the performances of the \href{https://mriqc.readthedocs.io/en/latest/}{MRIQC} pipeline on various large-scale datasets (ABIDE, $N=1102$ and CATI derived datasets, $N=9037$) used for both training and evaluation purposes. We focused our analysis on the MRIQC preprocessing steps and tested the pipeline with and without them. We further analyzed the site-wise and study-wise predicted classification probability distributions of the models without preprocessing trained on ABIDE and CATI data. Our main results were that a model using features extracted from \href{https://mriqc.readthedocs.io/en/latest/}{MRIQC} without preprocessing yielded the best results when trained and evaluated on large multi-center datasets with a heterogeneous population (an improvement of the ROC-AUC score on unseen data of 0.10 for the model trained on a subset of the CATI dataset). We concluded that a model trained with data from a heterogeneous population, such as the CATI dataset, provides the best scores on unseen data. In spite of the performance improvement, the generalization abilities of the models remain questionable when looking at the site-wise/study-wise probability predictions and the optimal classification threshold derived from them.

\section*{Introduction}
\label{intro}

Different effects can alter MRI data acquisition process and consequently image quality. They can be population-related or scanner-related, which we respectively refer to as study effect and site effect. Alteration of the acquisition process creates artifacts in the final image that can lead to errors in subsequent analyses. For instance, it has been shown that cortical thickness estimation varies with head movement severity \cite{alexander2016subtle, reuter2015head}. It is therefore crucial to control data quality right after acquisition to avoid any induced bias in the downstream analysis. 
\\ In the field of brain MRI quality control, the gold standard remains the expert's eye, in spite of several attempts to automatize the process \cite{mortamet2009automatic, alfaro2018image, esteban2017mriqc}. However, as the amount of MRI data available for analysis increases, it becomes necessary to rely on a robust automated quality control process that allows to evaluate large amounts of data. 
\newline
\\ Visual qualitative brain MRI quality control is time and resource consuming. Moreover, human rating is subjective, inducing variability between raters when inspecting the same data. A previous study showed that five trained raters who annotated 80 scans on a four-grade scale (from $0$ for bad data to $4$ for excellent data) had a mean inter-rater reliability of $0.53$ (scale from $0.0$ to $1.0$), ranging from $0.38$ to $0.72$ \cite{klapwijk2019qoala}. Automated quality control could yield more consistent ratings as well as rapid and standardized analysis of large amounts of data, even when originating from different studies and sites. 
\\ Several methods have been proposed in automated brain MRI quality control. 
\\ Early work proposed to derive two quality indices (\textit{$QI_1$} and \textit{$QI_2$}) from the background noise \cite{mortamet2009automatic}. The computation required two preprocessing steps: brain extraction and selection of voxels with artifact from the background. Both indices were tested on 749 1.5T and 3T images of the head in three-dimensional T1-weighted sequences from the Alzheimer’s Disease Neuroimaging Initiative (ADNI) database obtained from 36 sites. The results showed that the method made it possible to categorize artifacts on the basis of the scores given by the ADNI quality control center using a threshold determined \textit{a posteriori} on the values of the indices. However, this method would not detect artifacts altering exclusively the brain and not the background. For instance, some scanners or some acquisition sequences, such as MP2RAGE produce data with a very specific background, which makes this method unusable \cite{marques2010mp2rage}. In addition, the detection threshold strongly depends on the dataset. Last, only a small number of data with severe artifacts ($n=54$) was available in this study. 
\\ In more recent works, automated QC methods focused on applying machine learning classification methods on large sets of features, allowing the model to learn a mapping between the used features and the data quality.
\\ A fully automated QC classification pipeline, MRIQC \cite{esteban2017mriqc}, was proposed and tested to identify and correct an acquisition site effect using the ABIDE dataset for training (1102 subjects from 17 sites)\cite{di2014autism} and the DS030 dataset (265 subjects from 2 sites)\cite{gorgolewski2017preprocessed} for testing. A set of 69 features, based on the ones used in the Quality Assessment Protocol (QAP), were used \cite{shehzad2015preprocessed}. The results showed that, in spite of specific preprocessing steps (site scaling/centering, removing features correlated with the site, removing noisy features), an underlying site effect was still biasing the performances of the model. Another limitation of this work remains about the datasets that have been used. Even though the ABIDE dataset is multicentric, its population is rather specific (young and autistic subjects) and does not allow to clearly test the site effect on the held-out dataset (DS030), as this latter was acquired in only two different sites.
\\ A larger scale QC study, evaluating a group of 10,000 subjects from the UK Biobank dataset used heavy preprocessing methods for standardization and feature extraction which resulted in a total of 190 features \cite{esteban2017mriqc}. An overall classification model was then learnt by optimizing the false positive rate rather than the false negative rate. A more in-depth analysis was carried out in order to check the consistency of the scans of the first 6,000 subjects and the last 4,000 subjects. This analysis consisted in computing association tests between the features extracted by the pipeline and non-brain imaging variables and comparing the results between the two groups. The developed pipeline was very specific to the studied database without any generalization purpose. Indeed, as specified in the work, the pipeline was designed to limit visual QC as much as possible, hence the need to check for feature homogeneity in the data and optimizing the false positive rate.
\\ An alternative has been proposed in the Qoala-T pipeline: the QC was performed on FreeSurfer segmentation rather than raw data \cite{klapwijk2019qoala}. Several datasets were used, including the BrainTime dataset (784 scans from 3 time points), the Brain, Empathy, Social Decision-making dataset (BESD, 112 participants) and the ABIDE dataset (760 subjects from selected sites). QC was done on already segmented data from FreeSurfer, introducing biases from the different processing steps (non-brain tissue removal, cortical surface reconstruction, subcortical segmentation, cortical parcellation, estimation of various measures of brain morphometry) which could fail on bad quality data or pathological subjects. Correlation was found between the Qoala-T scores and the MRIQC metrics, hence the Qoala-T tool can serve as a proxy for raw image quality. An in-depth analysis showed that the most important measures for data quality prediction in their model were the left and right surface holes provided by FreeSurfer, which reflect topological defects in the surface reconstruction. However, the use of a segmentation pipeline to assess the quality of a data is clearly counter intuitive and quite a heavy computational cost, besides of potentially failing because of brain pathologies. When focusing on a scalable QC method, we seek for as little preprocessings as possible, unlike the one proposed by Qoala-T.
\newline
\\ In this work, we sought to evaluate the performances of an MRIQC classifier by evaluating its performance on unseen data when trained on different datasets. This software was chosen as it uses a set of characteristics commonly used in brain MRI analysis as features for QC classification. 
\\ Three datasets were used to test for various effects: 
\begin{description}
    \item[ABIDE] \cite{di2014autism}:  multi-center, homogeneous population (single study), medium size. 
    
    \item[DS030] \cite{gorgolewski2017preprocessed}: single center, homogeneous population (single study), small size. 
    
    \item[CATI]: multi-center, heterogeneous population (several studies), large size. 

\end{description}
Given the large scale of the CATI dataset ($9037$ scans), it was divided into different subsets to test relevant data characteristics (see section \ref{cati} for the details of the split). 
\\ Contributions of this work consists of :
\begin{enumerate}

    \item assessing the influence of the site effect (heterogeneity of acquisition sites) and the study effect (heterogeneity of population) on the performances of the MRIQC pipeline by using large heterogeneous sets of data for both training and testing. 
    
    \item testing the ability of a classification model to be robust to the site effect when removing all preprocessing steps.
    
    \item analyzing the site and study effects across datasets by looking at the site-wise and study-wise predicted classification probability distributions.
\end{enumerate}

We first describe the data and the different methods used in our work in the next section. Then, we show the results of our experiments before discussing them.

\section{Materials and methods}
\label{data_methods}
In this section, we will first describe the datasets used in our experiments (section \ref{datasets}) and then the methodological aspects of the QC pipeline and its analysis (section \ref{methods}). 

\subsection{Datasets}
\label{datasets}

The three datasets used in our experiments are described in this section: \href{http://fcon_1000.projects.nitrc.org/indi/abide/databases.html}{ABIDE}\cite{di2014autism}, \href{https://openneuro.org/datasets/ds000030/versions/00016}{DS030}\cite{gorgolewski2017preprocessed} and the CATI dataset. We defined the 'study' as a set of subjects from a particular population (depending on the pathology and/or the age) and the 'site' as a set of data acquired using the same MRI scanner with the exact same parameters (same sequences). See Table \ref{table1} for a summary of the presented datasets.

\begin{table}[htp]
\begin{adjustwidth}{-2.25in}{0in}
\footnotesize\setlength{\tabcolsep}{2.5pt}

\begin{tabular}{l@{\hspace{7pt}} *{8}{c}}
\hline
\textbf{Dataset} & \textbf{Size} &\textbf{\# artifact data} & \textbf{\# sites} & \textbf{\# studies} & \textbf{Mean population age} & \textbf{Pathologies} \\
\hline
\textbf{ABIDE} & $1102$ & $156$ $(14\%)$ & $17$ & $1$ & $17(\pm8)$& Autism \\ 
\textbf{DS030} & $265$ & $75$ $(28\%)$ & $2$ & $1$ & $33(\pm9)$ & Schizophrenia/bipolar disorder/ADHD \\
\textbf{CENIR} & $2189$ & $288$ $(13\%)$ & $3$ & $16$ & $18<mean<90$ & Various pathologies \\
\textbf{CATI\_train} & $4111$ & $537$ $(13\%)$ & $60$ & $35$ & $18<mean<90$ & Various pathologies \\
\textbf{CATI\_test} & $2737$ & $343$ $(13\%)$ & $60$ & $35$ & $18<mean<90$ & Various pathologies \\
\hline
\end{tabular}
\caption{Datasets used in this study and their characteristics}\label{table1}
\end{adjustwidth}
\end{table}

\subsubsection{\href{http://fcon_1000.projects.nitrc.org/indi/abide/databases.html}{ABIDE}}
\label{abide}

The Autism Brain Imaging Data Exchange (ABIDE) dataset contains 1102 subjects\cite{di2014autism}. The age of this population ranges from 6.5 to 64 years old with a mean of $17.0(\pm8.0)$ years.\\
For the manual QC, two different experts rated this dataset according to three labels (exclude/doubtful/accept). Each expert examined 601 MRI scans and a third expert reviewed the doubtful ones and set the final rating (see \cite{esteban2018improving, esteban2017mriqc} for more details). The number of exclude/doubtful/accept scans were 158 (14\%), 17 (2\%) and 928 (84\%) respectively.

\subsubsection{\href{https://openneuro.org/datasets/ds000030/versions/00016}{DS030}}
\label{ds030}

The DS030 dataset contains 265 scans from a study with healthy controls (130 subjects) and adults diagnosed with attention deficit/hyperactivity disorder (ADHD, 43 subjects),  bipolar disorder (49 subjects) or schizophrenia (50 subjects). The age range is 21 to 50 years old with a mean of $33(\pm9)$.
\\ For QC, the manual ratings provided by the work of \cite{esteban2017mriqc} were used. Please refer to their article for more details.  Using these ratings, the number of exclude/doubtful/accept scans were 75 (28\%), 145 (55\%) and 45 (17\%) respectively.

\subsubsection{CATI Dataset}
\label{cati}

The CATI dataset consists of 9,037 T1-weighted brain MR images from a large number of different sites and studies. The CATI (\textit{Centre pour l’Acquisition et Traitement des Images}) is a French national platform aiming to support multicenter neuroimaging studies\cite{operto2016cati}. This unique dataset is characterized by a significant heterogeneity of subjects, studies and sites. The best effort was made to standardize the acquisition parameters over all sites for studies fully managed by the CATI; some studies were initiated before the CATI and the acquisition protocol was not as closely matched between centers as in the CATI protocol, even though a standardization effort had been made to ensure reliable acquisitions. 
\\ All studies were conducted according to Good Clinical Practice guidelines. All participants provided written informed consent and the research received approval from a local ethical committee and regulatory agencies. The data used in this study were anonymized before access and analysis by the authors.
\\ For our experiments, the CATI dataset was divided into four subsets: 
\begin{itemize}
    \item \textit{CENIR dataset} is made of the data acquired at the CENIR (the neuroimaging analysis platform of the Paris Brain Institute, member of the CATI network), including 2,189 T1 weighted (T1w) scans, from three scanners and 16 studies.  
    
    \item \textit{CATI\_train dataset} is composed of 60\% of each study constituting the global CATI dataset excluding CENIR scans. This dataset included 4,111 T1w scans from 60 sites and 35 studies. 
    
    \item \textit{CATI\_sub dataset} is a subset of the \textit{CATI\_train dataset} composed of 2,189 T1w scans from 60 sites and 35 studies. It was constituted by randomly drawing data from each site from the \textit{CATI\_train} dataset in order to keep the same site distribution.
    
    \item \textit{CATI\_test dataset} is composed of the remaining 40\% of data per study and including 2,737 T1w scans from 60 sites and 33 studies. Two studies are missing from this dataset as they had too few scans to be included in both \textit{CATI\_train} and \textit{CATI\_test}. 
\end{itemize}

Scans from the CATI dataset were visually inspected by qualified experts using a standardized procedure and several quality control items were checked (among which wrinkles, blur, ghost, duplication, artifacts in localized areas such as hippocampus or cortex). Then a rater gave a global QC indicator, referred to as \textit{globalQualitative} with a $0-4$ scale, $0$ meaning very poor quality and $4$ meaning perfect quality.
\\ Data with $globalQualitative > 2$ were considered as qualitatively good.

\subsection{Methods}
\label{methods}

\subsubsection{The MRIQC Pipeline}
\label{mriqc_pipeline}
This study aims at evaluating the generalization performances of the MRIQC pipeline. The MRIQC framework version 0.16.1 was used for our whole work\cite{esteban2017mriqc}. 
\\The pipeline can be divided into three steps, which are described in more details below:
\begin{enumerate}
    \item Feature extraction, that aims at computing relevant characteristics using the raw image (section \ref{mriqc_preprocs});
    \item Preprocessing steps, that aim at normalizing the previously extracted features (section \ref{mriqc_feat_extraction});
    \item Model training and testing using the preprocessed features on a binary classification task (section \ref{model_training_methods}).
\end{enumerate}

\subsubsection{Feature Extraction}
\label{mriqc_preprocs}

The feature extraction consists of several steps (see \cite{esteban2017mriqc} for a detailed description) using different software (FSL\cite{jenkinson2012fsl}, ANTs\cite{avants2009advanced} and AFNI\cite{cox1997software}). From the skull stripping, spatial normalization and tissue segmentation outcomes, a set of 69 features (based on the Quality Assessment Protocol or QAP \cite{shehzad2015preprocessed}) was computed and used as input for classification. 

\subsubsection{Feature normalization}
\label{mriqc_feat_extraction}

In order to select the most effective preprocessing steps, a grid-search was used and tested for all possible configurations. The optimal configuration (best preprocessing steps and best classifier) was chosen by maximizing the ROC-AUC score. The corresponding preprocessing steps were the following:

\begin{itemize}
    
    \item The \textit{centering/scaling} steps which consist of a site-wise scaling and/or centering of a set of 36 chosen features (see SI \ref{S1_normed_feature_list} for the list of the features) to the first and third quantiles. Note that, at inference time, any data acquired in an unknown site (i.e. no scaling/centering parameters were estimated for this site during training) was scaled/centered with parameters estimated on the whole inference set. This is the default behavior implemented in \href{https://mriqc.readthedocs.io/en/latest/}{MRIQC}.
        
    \item The \textit{ft\_sites} step is a dimensionality reduction step based on extremely randomized trees that aims at excluding features highly predictive of the site\cite{geurts2006extremely}.

    \item The \textit{ft\_noise} step is a second feature selection aiming at discarding features providing less information than noise. This step is based on the Winnow algorithm using extremely randomized trees \cite{littlestone1988learning}. The features are compared to synthetically generated data (e.g. noise) and a signal-to-noise ratio (SNR) threshold is set, that measures how much information is embedded in the features when compared to noise. 
\end{itemize}

\subsubsection{Model training and testing}
\label{mriqc_classif}
The only model used in our experiments was the random forest classifier, as it was the best performing one according to \cite{esteban2017mriqc}. \\
Several hyper-parameters of this pipeline (whether to scale and/or center in the first step, the number of trees in the forest, their maximum depth,...) were chosen using a nested-cross-validation grid search scheme \cite{varoquaux2017assessing}. 
\\ A nested-cross-validation is a two-step cross-validation which starts by a classical train/test split of the data (outer loop) and then divides the training data into inner train/test sets (inner or nested loop). The inner split allows to select the best hyper-parameters and the outer split aims at assessing the performances of the model with the selected hyper-parameters. The final model is eventually refitted on the whole training dataset, using the selected preprocessing steps and model from the outer loop. 
\\To ensure that the cross-validation was not biased, two splitting schemes were chosen for the inner loop: 
\begin{itemize}
    \item \textit{Leave-One-Site-Out (LoSo)}: the data is split according to the acquisition site.
    \item \textit{KFold}: the data is randomly split into K folds.
\end{itemize}

The results were measured using the ROC-AUC score, as done in \cite{esteban2017mriqc}). After the entire cross-validation process was completed, the final model (learnt over the full training set) was evaluated on the other datasets.

\subsubsection{Performance/outcome analysis}
\label{mriqc_analysis_method}

\subsubsection*{Checking for a site effect with an unsupervised method}
\label{visualisation_clustering}
To assess the presence of a site effect in the different datasets, using the set of exatracted features, a two-dimensional t-distributed stochastic neighbor embedding (t-SNE) of the ten most represented sites per dataset was plotted (Figure \ref{fig1}). The t-SNE dimensionality reduction algorithm builds a representation where the distances between the different data points (i.e. feature vectors) are preserved at both local and global scale \cite{van2008visualizing}. Hence, the closest data in the original space (i.e. the space of the extracted features) are the closest ones in the projected two-dimensional space shown in figure \ref{fig1}.
\\ A quantitative measure was then computed by running a K-means algorithm over each dataset, using as many clusters as there were sites or studies (depending on the targeted effect). From each clustering, the completeness and homogeneity scores were computed and averaged over 1000 runs, the higher these scores (scale from $0.0$ to $1.0$), the more correlated the unsupervised clustering with the site/study repartition\cite{rosenberg2007v}. Note that the algorithm was run several times because of its undeterministic clustering due to the random initialization of the different partitions.

\subsubsection*{Model training and evaluation}
\label{model_training_methods}
The random forest classifier was trained on the four different training sets described above (ABIDE, CENIR, CATI\_train and CATI\_sub). The performances of the different learnt models were compared in the following experiments: \\
\begin{itemize}
    \item First, the classifier was trained on the various datasets using either a LoSo or KFold nested cross-validation scheme in the inner loop. This comparison provided information on the learning abilities of the models according to the number of scans, the variety of sites/studies in the learning data and the cross-validation splitting scheme.
    
    \item Second, the same training protocol was reiterated by removing the preprocessing steps (\textit{centering}, \textit{scaling}, \textit{ft\_sites} and \textit{ft\_noise}) and the results were compared to the previous ones. The aim was to test for ability of the model to learn to be robust to the site and study effect, without preprocessing the features. Moreover, given the different training datasets, the comparison between the pipelines with and without preprocessing provided better understanding of which characteristics of the training set improved generalization on unseen data.
    
    \item Third, we analyzed site-wise/study-wise prediction probability distributions of two models (learnt on ABIDE and CATI\_train). The optimal threshold of each distribution was calculated as the value maximizing the difference between the true positive rate (i.e. artifacted data classified as artifacted) and the false positive rate (i.e. non-artifacted data classified as artifacted), which gave a compromise between sensitivity and specificity. Then, the pairwise Wasserstein distances \cite{panaretos2019statistical} between intra-dataset sites/studies were reported in a matrix. 
    
\end{itemize}

By comparing the different site-wise (respectively study-wise) prediction distributions (shape, consistency) and the related thresholds, we were able to check for site (respectively study) specific biases on both training and evaluation data.

\section{Results}
\label{results}

Section \ref{visualisation} focuses on the quantification of the site and study effects in the different datasets. Sections \ref{all_preprocs_section} and \ref{no_preprocs_section} compare the ROC-AUC scores yielded by the MRIQC pipeline with and without preprocessing steps. Finally, section \ref{classif_thr_analysis} shows the probability distributions of the predictions given by two chosen models at a few sites/studies from ABIDE and CATI datasets.

\subsection{Checking for a site effect with an unsupervised method}
\label{visualisation}
\begin{figure}
    \noindent
    \includegraphics[scale=.4]{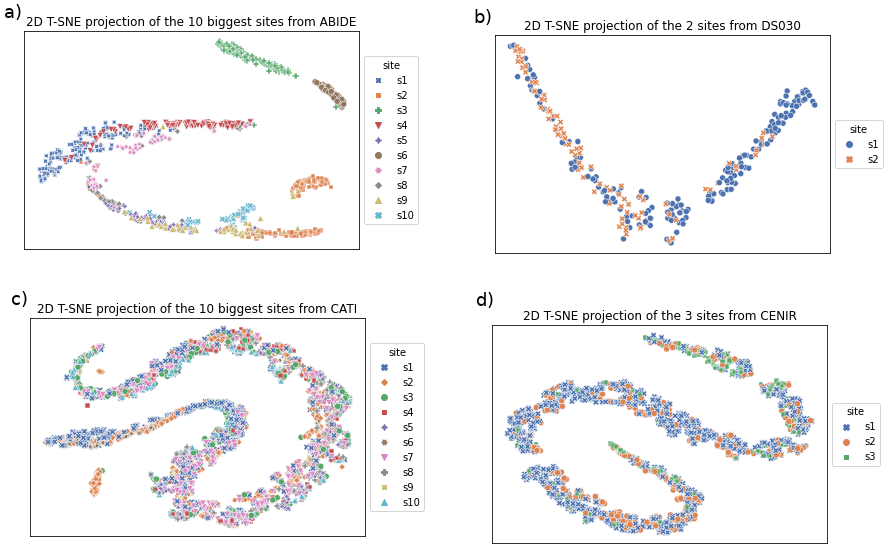}
    \caption{\label{fig1} 2D T-SNE visualisation of the data according to the acquisition site: \textbf{a)} ABIDE, \textbf{b)} DS030, \textbf{c)} CATI and \textbf{d)} CENIR. Features extracted from the different acquisition sites are displayed using different colors. A structured clustering of colors indicates a stronger site effect while a mixture of colors indicates a weaker site effect. Stronger site effect can be observed in the ABIDE dataset (upper left image) compared to the other datasets.}
\end{figure}

Figure \ref{fig1} shows a two-dimensional (2D) t-SNE projection of the extracted features colored according to the acquisition site. The ABIDE dataset presents a visually more structured clustering of the different sites/colors (upper left plot of the Figure \ref{fig1}). This suggests a stronger site effect in ABIDE compared to the other datasets.\\ 

The KMeans algorithm, run a thousand times, provided two quantitative measures (average of the completeness and homogeneity scores) of the correlation between the site distribution and the unsupervised clustering partitions (values in Table \ref{table2}). The two scores are much higher for the ABIDE dataset than the others, confirming a stronger site effect for this particular dataset. \\ 

\begin{table}[htp]
    \centering
    \begin{tabular}{|c|c|c|}
    \hline
    \bfseries \textbf{Dataset} & \textbf{Completeness score} & \textbf{Homogeneity score} \\
    \hline
    \bfseries ABIDE & $0.556 \pm 0.007$& $0.517 \pm 0.004$\\ 
    \bfseries CATI & $0.165 \pm 0.002$& $0.170 \pm 0.004$ \\
    \bfseries DS030 &$0.133 \pm 0.000$ & $0.143 \pm 0.000$\\
    \bfseries CENIR & $0.045 \pm 0.000$ & $0.055 \pm 0.000$ \\
    \hline    
    \end{tabular}
    \caption{\label{table2}Average completeness and homogeneity scores obtained after 1000 runs of K-means algorithm. The number of clusters $K$ was set as the number of different sites present in the datasets. The higher the scores, the more correlated the clusters and the sites repartition.}
\end{table}

For the study effect, the same experiment was repeated on the CATI and CENIR datasets only (ABIDE and DS030 include only one study). The average completeness and homogeneity scores were low for these two datasets indicating a weak bias due to the study effect (see Table \ref{table3}). \\

\begin{table}[htp]
    \centering
    \begin{tabular}{|c|c|c|}
    \hline
    \bfseries \textbf{Dataset} & \textbf{Completeness score} & \textbf{Homogeneity score} \\
    \hline
    \bfseries CATI & $0.143 \pm 0.001$ & $0.169 \pm 0.001$ \\
    \bfseries CENIR & $0.081 \pm 0.000$ & $0.109 \pm 0.001$ \\
    \hline    
    \end{tabular}
    \caption{\label{table3}Average completeness and homogeneity scores obtained after 1000 runs of K-means algorithm. The number of clusters $K$ was set as the number of different studies present in the datasets. The higher the scores, the more correlated the clusters and the studies repartition.}
\end{table}

\subsection{QC classification}
In the next two sections, results from two MRIQC pipelines (with or without the preprocessing steps) are presented.
\\ For each pipeline, four different training sets were tested: ABIDE, CENIR, CATI\_sub and CATI\_train. To measure the influence of the cross-validation splitting scheme, the models were trained using either a LoSo or a classical KFold scheme. 
\\ Section \ref{all_preprocs_section} analyzes the results yielded by the original pipeline\cite{esteban2017mriqc}.  
\\ Section \ref{no_preprocs_section} analyzes the results of a pipeline without preprocessing. 

\subsubsection{\textbf{Model with preprocessing}}
\label{all_preprocs_section}

The default pipeline provided by MRIQC was trained on four different datasets \cite{esteban2017mriqc}. The ROC-AUC scores obtained are shown in Table \ref{table4} and the preprocessing steps as well as the number of discarded features for each model are shown in Table \ref{table5}. \\

\begin{table}[htp]
    \begin{adjustwidth}{-2.25in}{0in}
    \begin{tabular}{|c*{7}{|c}|c|}
    \hline
    \backslashbox[6em]{\textbf{Test}}{\textbf{Train}}  & \multicolumn{2}{c|}{\textbf{ABIDE}} & \multicolumn{2}{c|}{\textbf{CENIR}} & \multicolumn{2}{c|}{\textbf{CATI sub}} & \multicolumn{2}{c|}{\textbf{CATI train}} \\
    \hline
    \textbf{Inner CV} & \textit{\textbf{LoSo}} & \textit{\textbf{5-Fold}} & \textit{\textbf{LoSo}} & \textit{\textbf{5-Fold}} & \textit{\textbf{LoSo}} & \textit{\textbf{5-Fold}} & \textit{\textbf{LoSo}} & \textit{\textbf{5-Fold}} \\
    \hline
    \textbf{Train cv} & \makecell{ $0.76$ \\($\pm 0.13$) }&
\makecell{$0.76$ \\($\pm 0.13$) }&
\makecell{$0.85$ \\($\pm 0.05$)} &
\makecell{$0.85$ \\($\pm 0.06$)} &
\makecell{$0.84$ \\($\pm 0.09$)} &
\makecell{$0.83$ \\($\pm 0.10$)} &
\makecell{$0.79$ \\($\pm 0.11$)} &
\makecell{$0.79$ \\($\pm 0.11$)}
\\ 
\hline
     \textbf{ABIDE} &  \textcolor{lightgray}{$0.96$} & 
\textcolor{lightgray}{$0.97$} &
$0.76$  & 
$0.62$  & 
$0.71$  &
$0.72$  &
$0.76$  &
$\mathbf{0.78}$ 
\\ 
\hline
 \textbf{DS030} &  $0.71$ &
$0.71$  &
$0.71$  &
$0.66$  &
$0.70$  &
$0.72$  &
$\mathbf{0.75}$ &
$0.75$  
\\
\hline
 \textbf{CENIR} & $0.74$ &
$0.75$ &
\textcolor{lightgray}{$0.94$} &
\textcolor{lightgray}{$0.93$} &
$0.80$  &
$0.80$  &
$\mathbf{0.81}$ &
$0.81$ 
\\
\hline
    \textbf{CATI test} &  $0.66$ &
$0.66$ &
$0.74$ & 
$0.73$ &
$0.80$ &
$0.79$ &
$\mathbf{0.81}$ &
$0.80$
\\
    \hline   
    \end{tabular}
    \caption{\label{table4} ROC-AUC scores of the models with all preprocessing steps. Values into brackets represent the standard deviation of the cross validation scheme ('Train cv' row). Terms in light gray are the results on the training dataset and in bold are the highest values per dataset (row).}
\end{adjustwidth}
\end{table}

\begin{table}[htp]
    \begin{adjustwidth}{-2.25in}{0in}
    \begin{tabular}{|c*{7}{|c}|c|}
    \hline
    \textbf{Train} & \multicolumn{2}{c|}{\textbf{ABIDE}} & \multicolumn{2}{c|}{\textbf{CENIR}} & \multicolumn{2}{c|}{\textbf{CATI sub}} & \multicolumn{2}{c|}{\textbf{CATI train}} \\
    \hline
     \textbf{Inner CV} & \textit{\textbf{LoSo}} & \textit{\textbf{5-Fold}} & \textit{\textbf{LoSo}} & \textit{\textbf{5-Fold}} & \textit{\textbf{LoSo}} & \textit{\textbf{5-Fold}} & \textit{\textbf{LoSo}} & \textit{\textbf{5-Fold}} \\
    \hline
    \textbf{Preprocs.} & [C.; Ft\_s.] & [C.; Ft\_n.] & [C.; Sc.] &  [Sc.; Ft\_s.; Ft\_n.] & [C.] & [C.] & [C.] & [Sc.] \\
    \hline
    \textbf{\# disc. ft.} &  [$0; 13$] & [$0; 20$] & [$0; 0$] & [$0; 2; 15$] & [$0$] & [$0$] & [$0$] & [$0$]\\
    \hline
    \textbf{\# disc. n. ft.} &  [$0; 11$] & [$0; 18$] & [$0; 0$] & [$0; 0; 8$] & [$0$] & [$0$] & [$0$] & [$0$]\\
    \hline
    \end{tabular}
    \caption{\label{table5} C.: site-wise centering; Sc: site-wise scaling; Ft\_s.: ft\_sites; Ft\_n.: ft\_noise. \\Preprocessing steps selected by the gridsearch (third row) for the pipeline with all preprocessings, with the total number of discarded features (fourth row) and the number of the discarded features which come from the normalized ones (fifth row).
    }
\end{adjustwidth}
\end{table}

\newpage

\subsubsection*{Cross-validation scheme (LoSo vs. 5-Fold)}

 The inner cross-validation scheme did not significantly alter the performances of the models trained on the ABIDE, CATI\_sub and CATI\_train datasets, which contain a large variety of acquisition sites. For the CENIR dataset (which only contains three different sites) a noticeable improvement with a LoSo split was seen, which reduced the overfitting of the model. By looking at the preprocessing steps selected by the pipeline and the number of discarded features from Table \ref{table5}, we observed that the CENIR-5Fold model discarded 17 features, amongst which 15 were assessed as noisy while the CENIR-LoSo model kept all the features. A possible explanation for these differences lies in the inaccuracy of the scaling step when using a 5-Fold split, which would not accurately estimate site-wise means and standard deviations. Note that the default behavior of the pipeline in inference when the site was unknown was to simply apply the selected normalization preprocessing (scaling or centering) to the inference data. In a LoSo scheme, given that the testing data belonged to the same site, the site-wise mean and standard-deviation estimations were accurate. 

\subsubsection*{ROC-AUC scores}
The highest ROC-AUC scores (in \textbf{bold}) were reached for the model learnt on the CATI\_train dataset (Table \ref{table4}). The score can be easily explained when evaluating on the CATI\_test as the same acquisition sites were present in both datasets (i.e. in training and evaluation), therefore the pipeline took advantage of the site specific scaling at evaluation time. The CATI\_train model also performed best on the unrelated datasets (ABIDE and DS030), which have both unseen acquisition sites and different manual quality annotation procedures.
\\ The results of the CATI and ABIDE datasets in particular were compared as they both present a large variety of acquisition sites allowing the evaluation of other multi-site unseen datasets. The DS030 was considered for consistency with the original MRIQC work, however given the size of the dataset, it was not looked in our analysis. The QC procedure trained on CATI data performed better when tested on ABIDE data than vice versa.   
\\ A comparison between CATI\_train and CATI\_sub provided a measure of the influence of the size of the training set on the performances. The ROC-AUC scores in cross-validation and in evaluation became more consistent as the amount of data increased. As expected from a cross-validation behavior, more data decreased overfitting (i.e. less differences between train and evaluation scores).  

\subsubsection*{Pipeline analysis}
Table \ref{table5} shows that scaled features were discarded in ABIDE and CENIR-5Fold. For the ABIDE-LoSo model, despite a proper site-wise centering estimation parameters (due to the site-wise splitting), 13 features were discarded, amongst which 11 were centered. Hence, we hypothesized that the centering did not manage to get rid of the site-effect found in section \ref{visualisation}. For both CENIR-5Fold and ABIDE-5Fold, the detrimental effect of the site-wise scaling/centering was accentuated by the fact that discarded normalized features were assessed as noise. 
\\To better understand the bias introduced by the preprocessing steps, the same experiment was repeated by removing these steps.

\subsubsection{\textbf{Model without preprocessings}}
\label{no_preprocs_section}

Next, the model training was reiterated by removing all preprocessing steps (\textit{centering}, \textit{scaling}, \textit{ft\_noise}, \textit{ft\_sites}). ROC-AUC scores of the pipelines are reported in Table \ref{table6}.

\begin{table}[htp]
    \begin{adjustwidth}{-2.25in}{0in}
    \begin{tabular}{*{8}{|c}|c|}
    \hline
    \backslashbox[6em]{\textbf{Test}}{\textbf{Train}}  & \multicolumn{2}{c|}{\textbf{ABIDE}} & \multicolumn{2}{c|}{\textbf{CENIR}} & \multicolumn{2}{c|}{\textbf{CATI sub}} & \multicolumn{2}{c|}{\textbf{CATI train}} \\
    \hline
    \textbf{Inner CV} & \textit{\textbf{LoSo}} & \textit{\textbf{5-Fold}} & \textit{\textbf{LoSo}} & \textit{\textbf{5-Fold}} & \textit{\textbf{LoSo}} & \textit{\textbf{5-Fold}} & \textit{\textbf{LoSo}} & \textit{\textbf{5-Fold}} \\
    \hline
    \textbf{Train cv} &  \makecell{$0.76$ \\($\pm 0.12$)} &
\makecell{$0.78$ \\($\pm 0.12$)} &
\makecell{$0.86$ \\($\pm 0.06$)} &
\makecell{$0.86$ \\($\pm 0.06$)} &
\makecell{$0.85$ \\($\pm 0.09$)} & 
\makecell{$0.85$ \\($\pm 0.09$)} &
\makecell{$0.84$ \\($\pm 0.10$)} &
\makecell{$0.84$ \\($\pm 0.10$)} 
\\ 
\hline
    \textbf{ABIDE} &  \textcolor{lightgray}{$0.97$} &
\textcolor{lightgray}{$0.97$} &
$0.77$ &
$0.78$ &
$\mathbf{0.82}$ &
$\mathbf{0.82}$ &
$0.79$ &
$0.79$
\\ 
    \hline
    \textbf{DS030} &  $0.72$ &
$0.71$ &
$0.76$ &
$0.75$ & 
$\mathbf{0.77}$ &
$\mathbf{0.77}$ &
$0.76$ &
$0.76$
\\
    \hline
    \textbf{CENIR} &  $0.79$ &
$0.78$ &
\textcolor{lightgray}{$0.96$} &
\textcolor{lightgray}{$0.96$} &
$\mathbf{0.85}$ &
$\mathbf{0.85}$ &
$\mathbf{0.85}$ &
$\mathbf{0.85}$
\\
    \hline
    \textbf{CATI test} &  $0.70$ &
$0.70$ &
$0.76$ &
$0.76$ &
$0.84$ &
$0.84$ &
$\mathbf{0.86}$ &
$0.85$
\\
    \hline    
    \end{tabular}
    \caption{\label{table6} ROC-AUC scores of the models without preprocessings. Values into brackets represent the standard deviation of the cross validation scheme ('Train cv' row). Terms in light gray are the results on the training dataset and in bold are the highest values per dataset (row).}
\end{adjustwidth}
\end{table}

\subsubsection*{LoSo vs. 5-Fold}
Again, the inner cross-validation scheme did not alter the results on ABIDE, CATI\_sub and CATI\_train datasets and no differences in scores were found between \textit{CENIR-LoSo} and \textit{CENIR-5Fold} either. This suggests that the removed steps (or at least one of them) reinforced the site effect, which was not significantly present, according to the results from section \ref{visualisation}. 

\subsubsection*{ROC-AUC scores}
For this pipeline, the highest scores (in \textbf{bold}) were reached by the \textit{CATI sub} model and surpassed the previous best results of the section \ref{table4}. This improvement can be explained by a notable decrease of overtraining induced by some of the removed preprocessings. 
\\The scores of the ABIDE model were either improved or unchanged compared to the previous pipeline. 
\\ The scores of all the CATI data based models were also improved when evaluated on the unrelated datasets (ABIDE and DS030). Again, removing the preprocessing steps allowed the model to make a better generalization on unseen data, even with datasets having a different annotation protocol. The results also confirmed that the most heterogeneous datasets in terms of acquisition sites and studies (CATI\_sub and CATI\_train) achieved better evaluation performances than a multi-center and mono-study dataset (ABIDE).
\\ Overall, this experiment showed that removing the preprocessing steps improved the scores.
\\ All the results shown above rely on the ROC-AUC score, which is a measure independent of any classification threshold. However, in practice, the use of a QC pipeline requires either a hard classification (good or bad data) or a reliable ranking (to limit the amount of data to visually control, for example). In the next section, focus is placed on analyzing the predicted classification probabilities of the models site-wise and study-wise.

\subsection{Predicted probabilities analysis}
\label{classif_thr_analysis}
\begin{figure}
    
    \hspace{-1.cm}
    \noindent   
    \includegraphics[scale=.4]{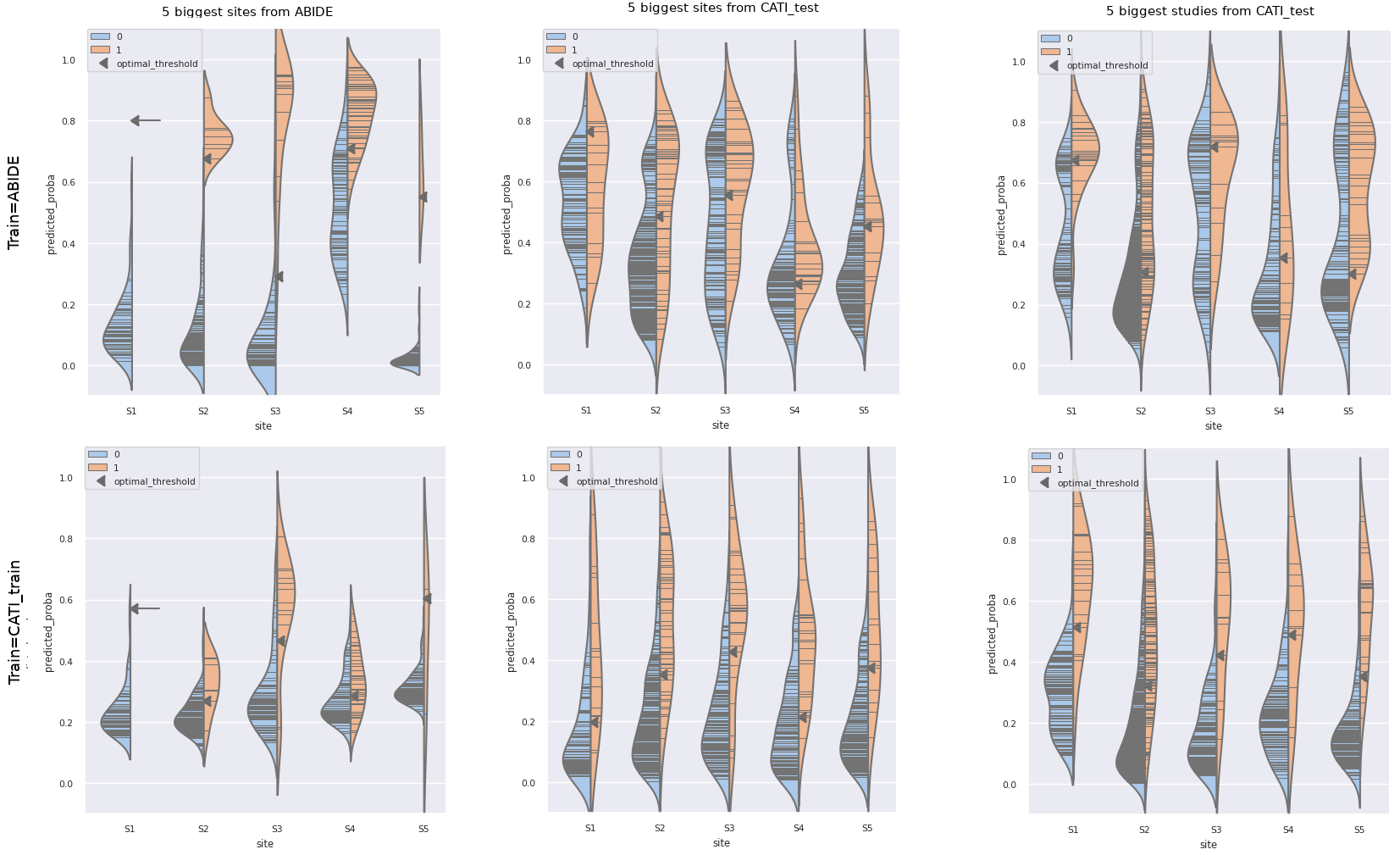}
    \caption{Predicted probabilities on the five biggest sites from ABIDE (left column), CATI\_test (middle column) and the five largest studies from CATI\_test (right column) by the models trained on ABIDE (upper row) and CATI\_train (lower row) without preprocessing. In blue non-artifacted data (class 0) and in orange artifacted data (class 1). Black lines inside the distribution represent the actual predicted values and the black arrows point at an optimal threshold computed as the maximal difference between the true positive rate (i.e. artifacted data classified as artifacted) and the false positive rate (i.e. non-artifacted data classified as artifacted).}
    \label{fig2}
\end{figure}

\begin{figure}
    
    \hspace{-1.cm}
    \noindent
    \includegraphics[scale=.4]{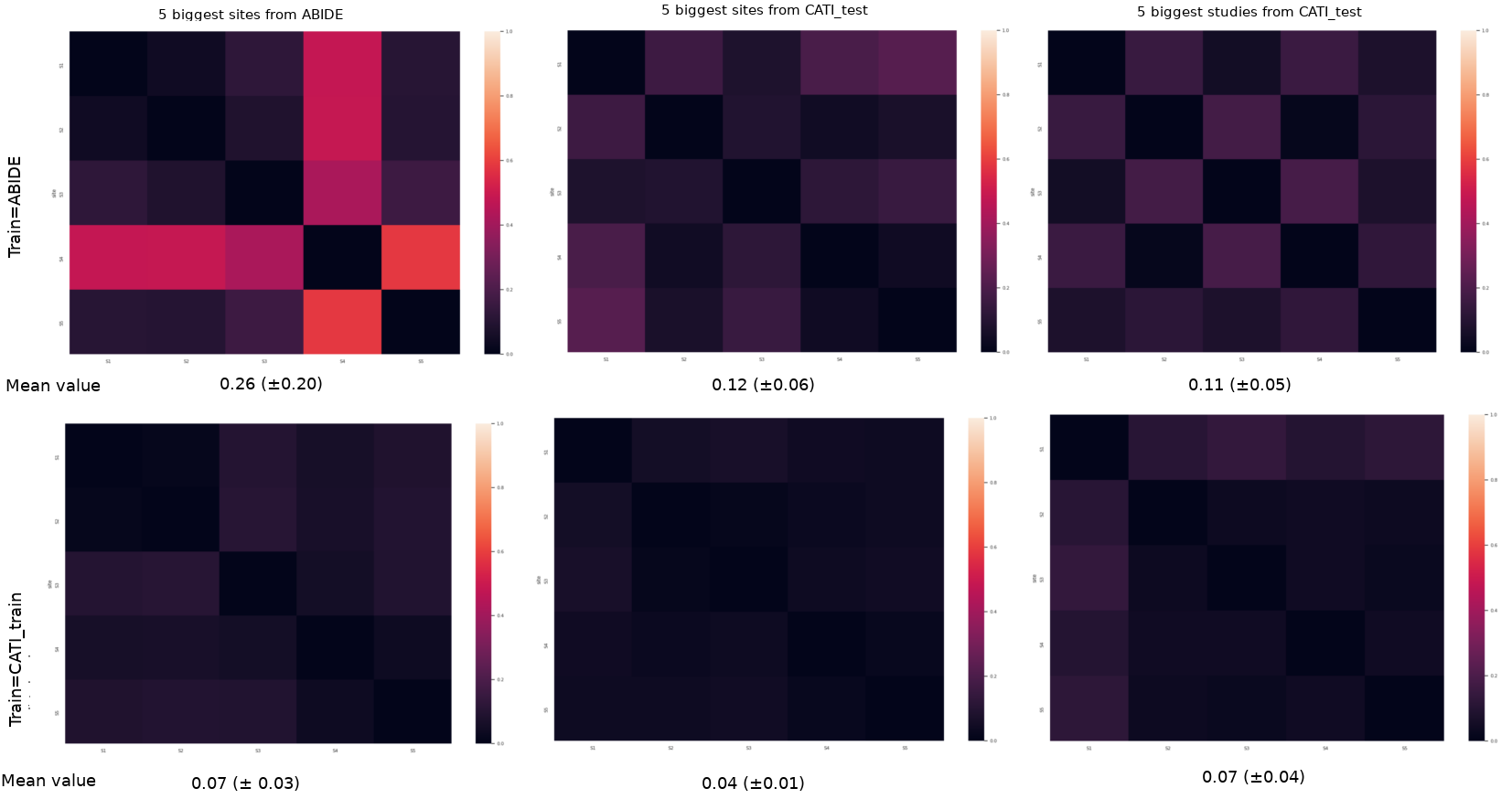}
    \caption{Pairwise Wasserstein distances between site wise predicted classification probabilities on the five biggest sites from ABIDE (left column), CATI\_test (middle column) and the five largest studies from CATI\_test (right column) by the models trained on ABIDE (upper row) and CATI\_train (lower row) without preprocessing. The mean value and the standard deviation of the site wise Wasserstein distances (computed on the upper triangular matrix without diagonal values) are written below each matrix.}
    \label{fig3}
\end{figure}

In this section, we focus on the classification probabilities predicted by the two models trained on ABIDE and CATI\_train without preprocessing and using a LoSo inner cross-validation scheme. 
\\ Figure \ref{fig2} shows the probability distributions of the predictions of the models trained on ABIDE and CATI\_train. The data were subdivided according to the classes: 0 in blue for good data (not artifacted) and 1 in orange for rejected data (artifacted). The distributions were plotted using the data from the five most represented sites from ABIDE and CATI\_test as well as the five largest studies from CATI\_test. An optimal threshold defined as the best compromise between specificity and sensitivity was also computed as the value maximizing the difference between the true positive rate (i.e. artifacted data classified as artifacted) and the false positive rate (i.e. non-artifacted data classified as artifacted).
\\ Figure \ref{fig3} shows the pairwise intra-dataset Wasserstein distances of the site/study wise predicted probability distributions of the distributions from the figure \ref{fig2} (the five most represented sites from ABIDE and CATI\_test as well as the five largest studies from CATI\_test). The mean values and the standard deviations of each pairwise distance matrix were reported as well. 

\newpage

The model trained on ABIDE (upper row) showed more irregular distribution shapes across sites/studies than the model trained on CATI\_train. For example, the optimal thresholds of the ABIDE model evaluated on CATI\_test ranged from $0.25$ for the site 4 to $0.8$ for the site 1. Consequently, inconstant and unpredictable performances were expected depending on the site or study targeted by the data, independently of the global ROC-AUC score ($0.70$) found in the previous section. The ABIDE model yielded bi-modal distributions (i.e. with two peaks) on the CATI data with, most of the times, one peak overlapping over the two classes. The bi-modal overlapping distributions suggested difficulties of the model to separate the classes using the learnt parameters. \\
On the other hand, the CATI\_train model predictions (lower row) looked more homogeneous, in spite of some variations of the optimal thresholds. The shapes of the predictions were mostly mono-modal distributions (i.e. with one peak), especially for good data (class 0, in blue), even when evaluated on ABIDE (unseen and independent dataset). Moreover, the range of the optimal thresholds were smaller (from $0.2$ to $0.6$) compared to the ones of the ABIDE model. 
\\ On figure \ref{fig3}, the Wasserstein distances from the site/study wise predictions of the model trained on ABIDE showed higher values (i.e. more variations in the distributions) than the ones from the model trained on CATI\_train. The discrepancy can be seen both visually (from the matrices) and quantitatively (from the mean values and standard deviation). For instance, the mean Wasserstein distance from the ABIDE trained model on ABIDE (the training set) was $0.26 (\pm 0.20)$ whereas the value from the CATI trained model on the same dataset was $0.07 (\pm 0.03)$.

\section{Discussion}

We have studied the performances of the MRIQC classification pipeline by first varying the learning data and then removing the preprocessing steps of the features prior to classification. We also analyzed the site-wise and study-wise classification probability distributions of two models without the preprocessing steps trained on large multi-site datasets. 
\\ Overall, we showed that the highest performances were obtained when learning with a multi-site and multi-study dataset (CATI\_train/CATI\_sub datasets), which provided better generalization on unseen data than learning with either multi-site/mono-study or mono-site/multi-study.
\\ Next, we showed that removing the site specific preprocessing steps (site-wise scaling and site-wise centering) improved the performances on unseen data. Hence, the model managed to be more robust to the site-effect without prior preprocessing of the features. 
\\ Last, the optimal thresholds and the per-class prediction probability distributions showed better class separability, i.e. better generalization when learning on a multi-study and multi-center dataset (the CATI\_train model compared to the ABIDE trained model), even when evaluated on a dataset with a strong site-effect.

\subsubsection*{Influence of the site/study effect on MRIQC}

The influence of biases inherent in batches of data (such as the site effect or the study effect) is known as the batch effect in the literature and needs to be correctly handled  to avoid confounding the outcome of interest \cite{leek2010tackling}. Prior to the evaluation of the MRIQC pipeline, we looked for site and study effects in the extracted features by using an unsupervised clustering method, which showed a stronger correlation between the site repartition and the clustering in the ABIDE dataset compared to the CATI based datasets. In spite of this batch effect, the pipeline yielded the best results on unseen data for the model trained on the CATI\_train dataset (ROC-AUC score of $0.78$ on ABIDE). Therefore, using a multi-site and multi-study dataset led to better generalization on an unseen dataset, even with a strong site-effect. 
\\ The manual QC assessment of the ABIDE and CATI datasets were done through different protocols, which might alter the scores of a model trained on one of them and evaluated on the other. Additionally, the populations from the ABIDE and the CATI dataset were different: the former included young autistic subjects while the latter included older subjects with pathologies. These differences between the two datasets made it more difficult to extrapolate the results from one to the other. In spite of that, the model learnt on the CATI data evaluated on ABIDE achieved a good ROC-AUC score of $0.78$. Moreover, this score was $0.12$ above the one obtained with the model learnt using ABIDE data and evaluated on CATI data.
\\ The site effect is a well known bias in brain MRI and has been shown to interfere with the outcome of analyzes such as Voxel Based Morphometry (VBM) or estimation of cortical volumes \cite{jovicich2006reliability, focke2011multi, radua2020increased, maikusa2021comparison}. However, to the best of our knowledge, this effect had never been tested in QC context with as large independent multi-center datasets as was done in our work. 

\subsubsection*{Removing the preprocessing steps}

By removing the preprocessing steps (\textit{site wise scaling}, \textit{site wise centering}, \textit{ft\_sites}, \textit{ft\_noise}) applied to the features prior to the classification, the results improved on an unseen multi-center dataset, showing the ability of the model to be more robust to the site effect. This counter-intuitive result went against the original purpose of the preprocessing steps, which was to improve the robustness of the model to the site effect \cite{esteban2017mriqc}. An improvement of the ROC-AUC score was seen for all models, even for the one trained on ABIDE, which presented the strongest site effect. Hence, the models managed to be more robust to the site effect. A possible explanation for this result could be that the preprocessing steps used to handle the batch effect were not appropriate and might have removed useful information from the features.
\\ Our conclusions from this experiment contradicted the benefit of using the preprocessing steps from the MRIQC pipeline. However, other methods, such as ComBat, exist in order to tackle batch effects and could be tested in order to assess their benefit in a QC task \cite{radua2020increased, li2021impact}. 

\subsubsection*{Site/study wise prediction distributions analysis}

To better assess and visualize the generalization performances of our models, the site wise/study wise prediction probabilities were represented along with the site wise/study optimal classification threshold (computed as the best compromise between specificity and sensitivity) and a quantitative value of the intra-dataset distances of the site/study wise predicted probabilities was computed using the Wasserstein distance. The distribution shapes were more consistent for the model trained on a larger and more heterogeneous dataset (CATI data versus ABIDE data), which allowed better performances on unseen sites from an independent dataset. This result was confirmed by the greater Wasserstein distances yielded by the model trained on ABIDE compared to the model trained on CATI\_train. Interestingly, the ABIDE trained model showed much greater Wasserstein distances on the training set, which suggests that the site-effect was actually learnt from the data and showed on the predictions.
\\ However, the inconstant value of the optimal classification threshold across sites/studies showed the limitation of the model on efficiently classifying data. Even though the model could be used for ranking, an undefined classification threshold would prevent it from classifying a single scanner at a time. Indeed, using the ROC-AUC score as a measure, which was independent of a classification threshold, did not reflect the classification performance but rather the ranking performance. Other measures should be used as well to assess the true classification performances of a model on single scanners pooled from different datasets.

\subsubsection*{Limitations of automated QC}

In automated brain MRI quality control, most methods rely on computationally expensive image high level preprocessing (skull stripping, spatial normalization, co-registration, segmentation) to compute features \cite{esteban2017mriqc, alfaro2018image, shehzad2015preprocessed, mortamet2009automatic}. In some cases, further preprocessing steps are applied to the features themselves to alleviate biases, such as the site effect \cite{radua2020increased}. Our results showed that by removing the feature preprocessing steps of the MRIQC pipeline and using the adequate dataset (CATI versus ABIDE), the results on unseen data were improved compared to the original ones from \cite{esteban2017mriqc} (an improvement of ~0.10 on the ROC-AUC score on unseen data for the model trained on CATI\_sub). However, we argue that better results may be achieved using different types of features. Previous work showed that the Euler number alone, computed by the FreeSurfer software, yielded better QC results than any of the features extracted by the QAP pipeline, on which the MRIQC features are based on \cite{rosen2018quantitative, dale1999cortical, shehzad2015preprocessed}. An obvious limitation of using the Euler number as a QC index is that it is based on a computationally expensive segmentation pipeline. Another risk in using high level processing methods is that these methods might fail with pathological subjects or even process heavily artifacted data without detecting anomalies.
\\ An automatic way to extract features of interest would be to learn them using deep learning models, using as little image preprocessing as possible. One major drawback from using neural networks on brain MRI data is the computation resources needed for the learning process, especially in the 3D case. Consequently, several works have focused on 2.5D (using several 2D slices), as in \cite{sujit2019automated}) or 3D patches, as in \cite{kustner2018automatic}.To the best of our knowledge, no work has truly used a large amount of data, which would allow to test for the learning abilities of the deep learning algorithms to learn features specific to MRI data in a QC context. 

\section{Conclusion}
The growing quantity of neuroimaging data and the need for efficient large-scale quality control make machine learning methods for MRI data particularly attractive. We have shown that machine learning algorithms are strongly influenced by their learning environment, including the training data and the different treatments applied to the data. By studying with the MRIQC pipeline, we concluded that using the adequate dataset not only improved the performance on unseen data but also allowed to remove all preprocessing steps originally aimed at mitigating site-effect biases. 
\\ However, in spite of this improvement, the method still had limitations including variations in the probability distributions of classifications per site and a computationally expensive end-to-end process. In future work, we will further explore methods relying on raw, or minimally processed MRI data, such as deep learning, to enable efficient large-scale processing of neuroimaging data.  

\section*{Supporting information}

\paragraph*{S1 Appendix.}
\label{S1_normed_feature_list}
{ \bf List of features normalized by the centering/scaling step: }
['cjv', 'cnr', 'efc', 'fber', 'fwhm\_avg', 'fwhm\_x', 'fwhm\_y', 'fwhm\_z',
    'snr\_csf', 'snr\_gm', 'snr\_total', 'snr\_wm', 'snrd\_csf', 'snrd\_gm', 'snrd\_total', 'snrd\_wm',
    'summary\_csf\_mad', 'summary\_csf\_mean', 'summary\_csf\_median',
    'summary\_csf\_p05', 'summary\_csf\_p95', 'summary\_csf\_stdv',
    'summary\_gm\_k', 'summary\_gm\_mad', 'summary\_gm\_mean', 'summary\_gm\_median',
    'summary\_gm\_p05', 'summary\_gm\_p95', 'summary\_gm\_stdv',
    'summary\_wm\_k', 'summary\_wm\_mad', 'summary\_wm\_mean', 'summary\_wm\_median',
    'summary\_wm\_p05', 'summary\_wm\_p95', 'summary\_wm\_stdv']
    
\section*{Acknowledgments}
This study was performed in the context of the CATI's methodological research program aiming at improving MRI acquisitions' quality assessment for a large set of different clinical research studies to allow efficient and reliable meta-analyses. 

\section*{Author contributions}
{\bf Data Curation:} Marie Chupin, Hugo Dary.

{\bf Methodology:} Ghiles Reguig, Romain Valabregue.

{\bf Software:} Ghiles Reguig, Romain Valabregue.

{\bf Supervision:} Romain Valabregue, Stéphane Lehéricy.

{\bf Validation:} Romain Valabregue, Marie Chupin.

{\bf Writing – Original Draft Preparation:} Ghiles Reguig.

{\bf Writing – Review $\And$ Editing:} Romain Valabregue, Marie Chupin, Stéphane Lehéricy, Hugo Dary, Eric Bardinet.
\nolinenumbers

%
%
%

\end{document}